\documentclass[10pt,twocolumn,letterpaper]{article}

\newif\ifcamera
\camerafalse

\usepackage[pagenumbers]{wacv} % To force page numbers, e.g. for an arXiv version

\usepackage{graphicx}
\usepackage{amsmath}
\usepackage{amssymb}
\usepackage{booktabs}

\usepackage[pagebackref,breaklinks,colorlinks]{hyperref}

\usepackage[capitalize]{cleveref} % TODO: not sure about this

\usepackage{import}
\usepackage{xcolor}
\usepackage{multirow}

\def\wacvPaperID{2103} % *** Enter the WACV Paper ID here
\def\confName{WACV}
\def\confYear{2025}

\begin{document}

%%%%%%%%% Custom macros

% Figue import and figure, section, equation, and table reference macros
\def\imgc#1#2#3#4{\def\svgwidth{#4}\import{figs/#1/}{#1.tex}\caption{#3}\label{fig:#2}}
\def\img#1#2#3{\def\svgwidth{\hsize}\import{figs/#1/}{#1.tex}\caption{#3}\label{fig:#2}}
\def\imgn#1#2#3#4{\def\svgwidth{\hsize}\import{figs/#1/}{#2.tex}\caption{#4}\label{fig:#3}}

\def\fg#1{Fig.~\ref{fig:#1}}
\def\sec#1{Section~\ref{sec:#1}}
\newcommand{\eq}[1] {Eq.~\ref{eq:#1}}
\newcommand{\tab}[1] {Table~\ref{tab:#1}}

% Editing commands
\definecolor{purple}{cmyk}{0.44, 0.92, 0, 0.01}
\newcommand{\DD}[1]{
  {\color{purple}#1}
}

\newcommand{\IM}[1]{
	{\color{blue}#1}
}
\definecolor{hiahia}{RGB}{255,0,128}
\newcommand{\YW}[1]{
	{\color{hiahia}#1}
}

\definecolor{gold}{rgb}{0.828, 0.683, 0.214}
\newcommand{\OT}[1]{
	{\color{gold}#1} % Because everything I suggest is pure GOLD!
}

\newcommand{\TODO}[1]{
	{\color{red}#1}
}

% Vector/Matrix commands
\newcommand{\R}{\mathbb{R}}
\newcommand{\vc}[1]{\mathbf{#1}}

\newcommand{\mytexttilde}{\raisebox{0.5ex}{\texttildelow}}
 % TODO: this has to be inside of the begin document?

% \teaser{
%     \img{teaser}{teaser}{
%         Subject (a) portrayed as a photorealistic 3D avatar (b-e) using our framework; as our framework allows for effortless integration with existing 3D pipelines and 3D assets, the avatar is imported in blender software and rendered with an additional 3D accessories, hair and glasses.
% }}

\title{3D Engine-ready Photorealistic Avatars via Dynamic Textures}

\author{Yifan Wang\\
Samsung Research America\\
{\tt\small y.wang1@samsung.com}
\and
Ivan Molodetskikh\\
Lomonosov Moscow State University\\
{\tt\small ivan.molodetskikh@graphics.cs.msu.ru}
\and
Ond\v{r}ej Texler\\
Samsung Research America\\
{\tt\small ondrej.texler@gmail.com}
\and
Dimitar Dinev\\
Samsung Research America\\
{\tt\small dimitar.d@samsung.com}
}

%%% The darkest magic to have the Teaser on the 1st page

%% Comment this if the below code block is not commented 
%\maketitle
%% Uncomment this and comment the \maketitle above
\twocolumn[{%
	\renewcommand\twocolumn[1][]{#1}%
	\maketitle
	\begin{center}
		\centering
		\captionsetup{type=figure}
			\def\svgwidth{\hsize}%% Creator: Inkscape 1.3.2 (091e20e, 2023-11-25, custom), www.inkscape.org
%% Accompanies image file 'teaser.pdf' (pdf, eps, ps)
\begingroup%
  \makeatletter%
  \providecommand\color[2][]{%
    \errmessage{(Inkscape) Color is used for the text in Inkscape, but the package 'color.sty' is not loaded}%
    \renewcommand\color[2][]{}%
  }%
  \providecommand\transparent[1]{%
    \errmessage{(Inkscape) Transparency is used (non-zero) for the text in Inkscape, but the package 'transparent.sty' is not loaded}%
    \renewcommand\transparent[1]{}%
  }%
  \providecommand\rotatebox[2]{#2}%
  \newcommand*\fsize{\dimexpr\f@size pt\relax}%
  \newcommand*\lineheight[1]{\fontsize{\fsize}{#1\fsize}\selectfont}%
  \ifx\svgwidth\undefined%
    \setlength{\unitlength}{1518bp}%
    \ifx\svgscale\undefined%
      \relax%
    \else%
      \setlength{\unitlength}{\unitlength * \real{\svgscale}}%
    \fi%
  \else%
    \setlength{\unitlength}{\svgwidth}%
  \fi%
  \global\let\svgwidth\undefined%
  \global\let\svgscale\undefined%
  \makeatother%
  \begin{picture}(1,0.25296443)%
    \lineheight{1}%
    \setlength\tabcolsep{0pt}%
    \put(0,0){\includegraphics[width=\unitlength,page=1]{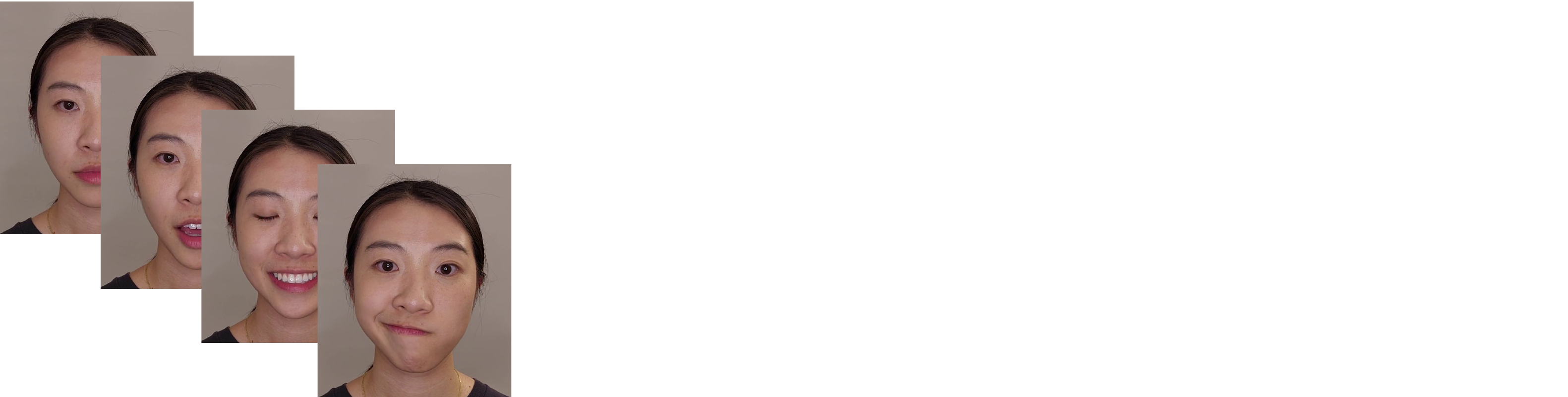}}%
    \put(0.01645876,0.23008273){\color[rgb]{0,0,0}\makebox(0,0)[t]{\lineheight{0}\smash{\begin{tabular}[t]{c}(a)\end{tabular}}}}%
    \put(0,0){\includegraphics[width=\unitlength,page=2]{teaser_x.pdf}}%
    \put(0.34649829,0.23008276){\color[rgb]{0,0,0}\makebox(0,0)[t]{\lineheight{0}\smash{\begin{tabular}[t]{c}(b)\end{tabular}}}}%
    \put(0.51052991,0.23008276){\color[rgb]{0,0,0}\makebox(0,0)[t]{\lineheight{0}\smash{\begin{tabular}[t]{c}(c)\end{tabular}}}}%
    \put(0.68147852,0.23008276){\color[rgb]{0,0,0}\makebox(0,0)[t]{\lineheight{0}\smash{\begin{tabular}[t]{c}(d)\end{tabular}}}}%
    \put(0.851439,0.2300827){\color[rgb]{0,0,0}\makebox(0,0)[t]{\lineheight{0}\smash{\begin{tabular}[t]{c}(e)\end{tabular}}}}%
  \end{picture}%
\endgroup%
\caption{
			Subject (a) portrayed as a photorealistic 3D avatar (b-e) using our framework; as our framework allows for effortless integration with existing 3D pipelines and 3D assets, the avatar is imported into Blender software and rendered with additional 3D accessories. %, hair and glasses.
	}\label{fig:teaser}
	\end{center}
}]

%%%%%%%%% ABSTRACT
\begin{abstract}
%There is an emerging trend to generate photorealistic and animatable avatars in the current AI-generated-content field. Existing methods suffer deficiencies from different aspects that prevent them from being comfortably used in real-world scenarios by ordinary customers. Such as the requirement of commercial studio capture system and heavy computation resources, the use of implicit representation (e.g., voxels used in NeRFs) that is hard for game engine integration, or the lack of key facial details like teeth and tongue textures. In this work, we propose an end-to-end pipeline that builds explicitly-represented photorealistic 3D avatars, an effective yet efficient method that achieves comparable visual quality to the state-of-the-art 3D avatar generation while allowing for seamless integration into existing graphics pipelines. 
As the digital and physical worlds become more intertwined, there has been a lot of interest in digital avatars 
that closely resemble their real-world counterparts. Current digitization methods used in 3D production pipelines 
require costly capture setups, making them impractical for mass usage among common consumers. Recent academic 
literature has found success in reconstructing humans from limited data using implicit representations (e.g.,
voxels used in NeRFs), which are able to produce impressive videos. However, these methods are incompatible with 
traditional rendering pipelines, making it difficult to use them in applications such as games. In this work,
we propose an end-to-end pipeline that builds explicitly-represented photorealistic 3D avatars using standard
3D assets. Our key idea is the use of dynamically-generated textures to enhance the realism and visually mask 
deficiencies in the underlying mesh geometry. This allows for seamless integration with current graphics
pipelines while achieving comparable visual quality to state-of-the-art 3D avatar generation methods.

%\keywords
%photorealistic avatars,  dynamic textures, 3D reconstruction

\end{abstract}

%%%%%%%%% BODY TEXT
\section{Introduction}
\label{sec:intro}
Digital reconstruction of humans is a rapidly growing area in both computer vision research and in the broader
industry. The creation of photo-real digital avatars enables rapid generation of novel videos of a particular
subject, which has found many use-cases in the services sector such as online education, advertising, hospitality, 
customer service, and in telepresence. The current paradigm for creating a digital avatar typically involves hiring
an actor and recording their performance in a capture studio. While the cost and hardware requirements for such a
capture are low when compared to the elaborate Hollywood studios used for summer blockbusters, they are still
prohibitively expensive for common people. In the entertainment industry, there has been exponentially increasing 
demand for personalized avatars for use in applications such as games or VR-assisted meetings. Companies like Meta 
Platforms Inc and Roblox have introduced platforms and ecosystems where users can interact with each other digitally,
using avatars that roughly represent themselves using intentionally-simplified art styles and a combination of
pre-defined facial features, akin to a character creator commonly found in video games. While these can sometimes 
approximate an individual, the general problem of realism and, more importantly, recognizability remains.

Traditional approaches to creating characters involve two main components: the head geometry model and
the textures. The head geometry is usually represented with a parametric model typically using 
blendshapes~\cite{lewis2014practice}. Actors are brought in, scanned, and fitted to the models. During this process,
high-quality textures of many varieties are produced (e.g., albedo, specular maps, or bump maps) to capture
the appearance as accurately as possible. Once these assets have been created, they remain mostly static
and are consumed by the 3D engine. A skilled team of artists can very accurately reconstruct a
person's identity, and these techniques are already commonly used in modern film and game development. The big
drawback to these approaches is the intense manual labor required to create high-quality assets and 
produce good results.

In the modern academic literature, there has been a rapid increase in works that seek to automatically and 
accurately reconstruct humans without requiring intensive manual labor; those techniques usually 
leverage the power of neural rendering techniques ranging from convolutional variational 
auto-encoders~\cite{Lombardi18} and volumetric techniques~\cite{Lombardi19,Lombardi21,cao22,nehvi2023360deg}
to even diffusion-based models~\cite{stypulkowski23,shen23,he23}.
While on the surface these techniques produce very accurate
and recognizable avatars, with the exception of~\cite{Lombardi18}, they rely on screen-space neural rendering to 
achieve the promised level of quality. While this is not an issue when producing video content, it becomes an issue when one
wants to use these avatars in an interactive application like an online social video game. Screen-space rendering 
is an alternative to the traditional z-buffer rasterization pipeline used in most modern 3D game engines, and while 
one can employ tricks such as bill-boarding to integrate the two, this integration is far from perfect, is not practical, 
and comes with many limitations.

In this work, we will focus on creating output that is compatible with 3D environments, using meshes and dynamic 
textures similar to~\cite{Lombardi18}. While this work required an expensive multi-view capture setup with carefully
tracked and fitted meshes, we will limit our data and training requirements to consumer-grade hardware to 
further increase the practicality of the proposed method. While a full body would be desirable for certain applications,
the scope of this work will be limited to only the face, as this is the most important for recognizability.
Our method consists of two main components: the reconstruction and the rendering.

{\bf The reconstruction module} aims to fit the user's likeness to our underlying head model and to extract all 
of the relevant parameters for the short capture sequence. We opted for the FLAME 3D model~\cite{Li17_FLAME} 
as it is a widely used model with an ecosystem of frameworks and tools around it created by the research community.
A typical challenge for a reconstruction process is the entanglement of the identity (i.e., intrinsic geometric parameters) 
and expression (i.e., dynamic parameters). In order to accurately disentangle the two, we propose to split the 
capture into two parts. 
%
%The first video focuses on creating an accurate 3D mesh reconstruction with 
%a neutral face expression which we then fit only the identity parameters of 
%our model to. The second video aims to extract the views and expressions we
%need to train our rendering network, and contains a short scripted sequence 
%for people to record themselves reenacting.
The first part focuses on creating an accurate 3D mesh reconstruction. It requires a short 
(roughly $20$ seconds) video clip consisting of different angles of the user with a neutral face expression; we use it to fit only the identity parameters of our model.
The second part aims to capture different viewpoints and expressions that we will need for 
training our rendering network. This requires the user to record themselves reenacting a 
short (roughly $60$ seconds) scripted sequence of facial expressions and speech from a fixed single camera view.
This provides training data for the rendering module.

{\bf The rendering module} consists of training our small convolutional rendering network
on the aforementioned small dataset. To ensure practicality, it is vital to 
keep the model size low and the inference speed fast. We accomplish this by first extracting a 
high-quality static texture and subsequently applying a neural network to fix various errors 
that might be present in the texture and supplement the missing detailed texture. Having a 
static-texture as a starting point requires significantly 
less capacity from the network when compared to a setup where a Variational Autoencoder, 
Generative Adversarial Network, or Latent Diffusion based approach would be used to generate
an entire texture from scratch.
We show that even with such a small dataset and limited hardware, we are able to reconstruct recognizable
faces that can  be used in 3D engines. Our rendering quality in the face area is also comparable to current
state-of-the-art screen-space renderers. When loaded in a 3D engine, our head can be enhanced in all of the
standard ways such as adding accessories like glasses or changing the hair style. 

In summary, our contributions are:
\begin{itemize} 
	\item A data capture process focusing on extracting the maximum amount of necessary information 
          while keeping the capture requirements minimal.
	\item A two-stage identity and expression extraction pipeline that disentangles the two, 
          obtaining a better fitted mesh than current monocular methods.
	\item An optimized convolutional renderer focusing on inference speed and model size.
\end{itemize}

\section{Related Work}
\label{sec:related}
%{\bf Facial Avatar Image/Video Synthesizing}
%The well-developed generative adversarial networks--GANs~\cite{Goodfellow14}, 
%in the past a few years have enabled significant advancements in synthesizing images and videos. 
In the past decade, a significant advancement in creating images and videos 
has been enabled by generative adversarial networks--GANs~\cite{Goodfellow14}.
Naturally, a lot of those techniques, often dubbed ``deepfakes'', focus on synthesis of human
faces~\cite{Suwajanakorn2017,Vougioukas19,Thies2016face2face,Bregler1997,Zakharov2019FewShotAL} 
with a particular emphasis on photorealism. Some methods to generate animations of 
talking faces are based on re-enactment usually driven by a target
video~\cite{hong2022depth,wang2021facevid2vid,wayne2018reenactgan,OneShotFace2019,nirkin2019fsgan,nirkin2022fsganv2}, other methods are driven by 
speech~\cite{song2018talking,zhou2019talking,zhang2022text2video,Yang2020_MakeItTalk,Prajwal2020lip}; as speech, (a form of audio), can be represented using a large variety of different 
intermediate representations, some methods chose to leverage 2D
keypoints~\cite{Suwajanakorn2017,Chen2019_hierarchical} while others proposed to use meshes~\cite{nagano2018pagan, Cudeiro2019capture,Thies2020neural,Richard_2021_ICCV} or hybrid approaches 
using both driving-keypoints and audio features~\cite{Ravichandran23-CVPR}.
However, most of those techniques do not consider consistency across multiple views 
and are often limited to only front-facing view-points.

%%%%%%%%%%%%%%%%%%%%%%% Moved here from the method section for a better figure placement
\begin{figure*}[ht]
\def\svgwidth{\hsize}\import{figs/pipeline_fitting/}{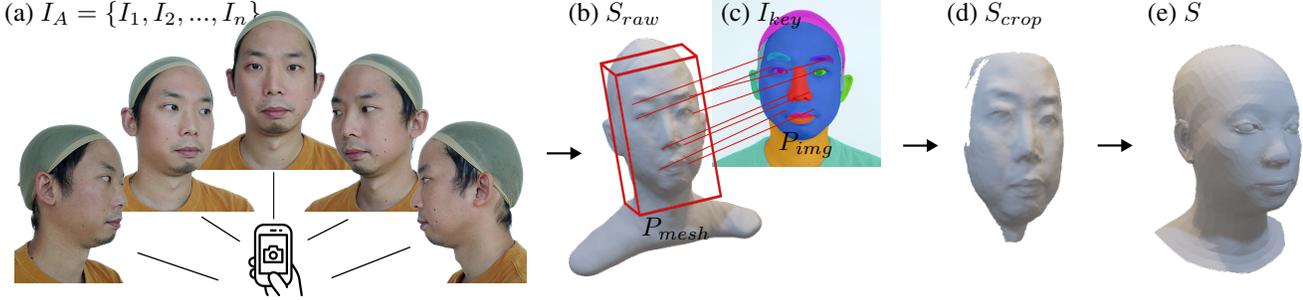}\caption{
A precisely fitted mesh is required for our subsequent rendering pipeline. 
To reconstruct a mesh, the subject 
(a) captures themselves using a mobile phone, 
(b) Neural Deferred Shading (NDS) is used to reconstruct the  mesh $S_{raw}$,
a face portion $S_{crop}$ (d) of $S_{raw}$ is extracted via a bounding box determined by $P_{mesh}$ which is back-projected from 
$P_{img}$ in $I_{key}$ (c), 
and finally a 3DMM FLAME mesh is fitted (e) to the face portion $S_{crop}$.
}\label{fig:pipeline_fitting}
\end{figure*}

{\bf Diffusion-based Techniques.}
Denoising diffusion probabilistic model~\cite{ho20} and later denoising 
diffusion implicit model~\cite{song21} based techniques have dominated the image and 
video generative space~\cite{dhariwal21,ramesh21} and they were made 
substantially more practical by delegating computationally intensive training and 
inference into a lower-dimensional latent space~\cite{esser20,rombach21}. 
Naturally, diffusion based methods started to also be explored for the task of 
face animation and talking head 
generation~\cite{stypulkowski23,shen23,du23,zhua23,zeng23,he23}.
However, the mentioned diffusion models do not leverage any 3D priors making them 
impractical in use-cases where there is a need for a very precise control over the 
view-point, head rotation, or face expressions.

{\bf Neural Volume-based Facial Avatars.}
To ensure the cross-view geometric consistency, many methods successfully leveraged differentiable ray-marching and 3D volume representations to create talking and animatable avatars. Neural volumes~\cite{Lombardi19} first represents an avatar as a dynamic 3D volume and uses a dynamic warping field to overcome the limitation of voxelization resolution. Mixture of Volumetric Primitives~\cite{Lombardi21} further improves the efficiency of the volumetric approach by only ray-marching on a set of small key volumes. With the great boosting of neural radiance fields, ~\cite{park2021nerfies,park2021hypernerf,hong2022headnerf,mihajlovic2022keypointnerf} demonstrate the success of the static avatar novel view synthesizing. Nerface~\cite{gafni2021dynamic}, IMAvatar~\cite{Zheng22_imavatar}, and Instant Avatar~\cite{zielonka2023instant} learn the dynamic, implicit neural avatar from only monocular videos with 3D Morphable Models (3DMM)~\cite{Blanz1999AMM,Knothe2011} as the proxy for controllability or deformation reference. Instant Avatar also greatly improves the generation speed by incorporating~\cite{muller2022instant}. Otavatar~\cite{ma2023otavatar} and~\cite{li2023generalizable} enable one-shot animatable 3D avatars. Beside of 3DMM-based controllability when animating neural implicit avatars, AD-NeRF~\cite{guo2021ad} drives the neural implicit talking head by audio signals. LatentAvatar~\cite{xu2023latentavatar} fully gets rid of the 3DMM dependency and drives the neural head avatar by latent expression code learned in a self-supervised manner.
In order to synthesize 3D-aware avatar video that improves cross-view geometry 
consistency, \cite{shi2021lifting} adapts 2D StyleGAN to predict both images and depth maps. 
StyleNeRF~\cite{gu2021stylenerf} and CIPS-3D~\cite{zhou2021cips} both inject 3D 
information by integrating small NeRF~\cite{Mildenhall2020nerf} components into 
their style-based generators. EG3D~\cite{chan2022efficient} proposes an efficient 
tri-plane 3D representation to lift the information from images to 3D space. 
Despite the huge momentum of NeRF-based virtual avatar generation, how to integrate this 
representation into existing 3D pipelines such as 3D game engines (e.g., Unity or Unreal Engine) 
still remains under-explored, which restricts NeRF-based avatar applications still to be image-based.

% The section above already mentiones some 3DMM based approaches
% we should perhaps move some 3DMM citations from the above to this section
{\bf 3DMM-based Facial Avatars.}
Even in neural volume-based avatar generation, we can see the heavy reliance on 3DMM. 3DMM demonstrates its great ability representing the shape and motion of human faces in~\cite{cao2016real, Li17_FLAME,egger20203d,blanz2023morphable,li2020learning, hu2017avatar}. 3DMM represents the shape, expression and texture of a person by linearly combining a set of bases using person-specific parameters. Thanks to the differentiable rendering techniques, fine-grained avatar facial texture learning from unbounded images becomes quite effective~\cite{luo2021normalized}, and numerous methods~\cite{feng2021learning,danvevcek2022emoca,filntisis2023spectre,dib2021towards} have been proposed to achieve dynamic 3DMM model alignment, tracking and animation in self-supervised manner from in-the-wild images and videos. These methods pave the way to a series of neural-textured facial avatar generation with explicit representation. NHA~\cite{Grassal22_neural_head_avatars} learns to deform the per-frame roughly tracked 3DMM FLAME~\cite{Li17_FLAME} mesh to get the fine-grained shape fitting and learn to optimize the neural texture jointly. The deformation directly operates on vertices and is prone to non-manifold issue. ROME~\cite{Khakhulin2022ROME} uses the per-frame FLAME fitting from~\cite{feng2021learning} as posed geometry prior and focuses more on deforming the hair shape for better avatar hair rendering with more accurate head shape. However, fitting a 3DMM according to monocular images is ill-posed due to the occlusions of multi views. On top of that, methods~\cite{feng2021learning,danvevcek2022emoca,filntisis2023spectre,dib2021towards} try to solve expressions, cameras, identity, lights, poses, etc. in one go, which makes their model distracted and results in a relatively averaged face. MICA~\cite{zielonka2022towards} uses a medium-scale 3D dataset and a pre-trained face recognition network to improve the uniqueness of 3DMM identity recovery. \cite{giebenhain2023learning} proposes to reconstruct a neural parametric model and can disentangle shape and expression control, but the input has to be in a 3D format. 
Our method aims for explicit 3D avatar generation and seeks to help traditional 3D modellers and artists to integrate such avatars into a game engine without pain.
%, \cite{grassal2022neural} what is this citation here? 

\section{Our Approach}
\label{sec:method}

\begin{figure}
    \centering
    \includegraphics[width=0.48\textwidth]{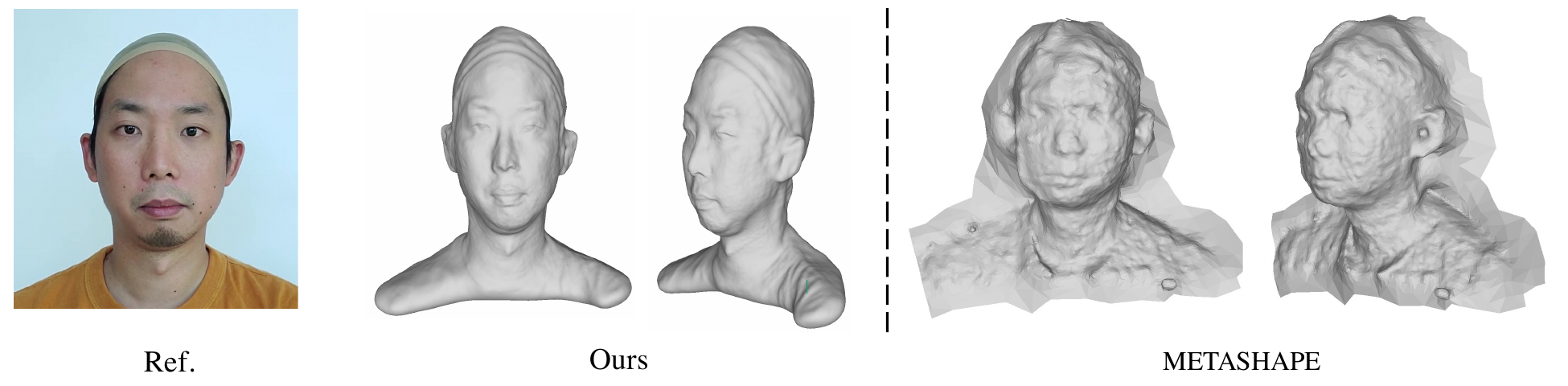}
    \caption{A reconstruction comparison between our adapted NDS and Metashape using the same set of COLMAP-calibrated images (under 20) extracted from Clip-A.}
    \label{fig:NDS_vs_metashape}
\end{figure}

Our key statement is that the high-fidelity recognizable dynamic 3D avatars are achievable by leveraging
a combination of classical methods, which employ meshes and textures, and the idea of dynamic 
textures~\cite{Lombardi18} that are rendered on-the-fly using a lightweight (convolutional) network.
Furthermore, we will restrict ourselves to a simple data capture pipeline using only monocular video sequences 
captured using common modern cell phones. Our desired video should contain two main parts: one (Clip-A) for determining 
the identity and another (Clip-B) for dynamic texture acquisition (see details in the appendix).

% In our capture, Clip-A 
% is a casual multi-view capture, typically under $20$ seconds and Clip-B captures the frontal view 
% expressions and talking, which is around $60$ seconds.

\subsection{Identity-preserved Mesh Reconstruction}
\label{sec:recons}
Our 3DMM-based dynamic avatar generation pipeline first recovers subject's immutable 
identity information from Clip-A. While there exist many 
works~\cite{feng2021learning,danvevcek2022emoca,zielonka2022towards}
that also extract model parameters from monocular images, these usually do not provide desired accuracy, and output highly
regularized results due to the under-constrained nature of monocular input. Moreover, some entangle the expression and
identity parameters, leading to further inaccuracies. 
%The aforementioned identity video consists of

To accurately capture the identity, Clip-A needs the user to capture themselves from multiple angles with minimal movement and neutral facial expression. More details on the capture setup are given in the appendix. In Clip-A we sample a set of key frames 
$\textbf{I}_A = \{I_1, I_2, ..., I_n\}$ (\fg{pipeline_fitting}a), from which we estimate the camera parameters using COLMAP ~\cite{schoenberger2016sfm}.  

Given this set of calibrated images $\textbf{I}_A$, we adapt and improve the optimization stage in Neural Deferred Shading
(NDS)~\cite{worchel2022multi} such that it reconstructs a high-fidelity mesh $S_{raw}$ (\fg{pipeline_fitting}b) with facial 
detail. NDS is an object contour-based mesh deformation optimization starting from coarse visual hull and iteratively 
adding high-frequency details via its jointly-optimized neural texture module. NDS offers considerable advantages over 
classic photogrammetry reconstruction (e.g., the commercial software Metashape)
%\footnote{https://www.agisoft.com/}
which requires substantially more multi-view photos and usually outputs significantly noisier mesh reconstruction; NDS's 
mesh reconstruction is watertight, smooth, and conveys adequate high-frequency facial features for the subsequent 3DMM model 
registration. An illustration example can be found in Fig. \ref{fig:NDS_vs_metashape}.

We chose the FLAME model~\cite{Li17_FLAME} as our 3DMM to fit $S_{raw}$ due to its widespread usage in 3D 
computer graphics pipelines. Since we are most concerned with the face and hope the 3DMM fitting only focuses on the face region 
without overfitting to, e.g., the hair and the neck, we need to crop $S_{raw}$ such that the resulting mesh only contains the face part. 
To do so, we first select a frontal-view key frame $I_{key}$ from $\textbf{I}_A$ and detect 2D facial landmarks 
$P_{img} \in \mathbb{R}^{N \times 2}$ using~\cite{lugaresi2019mediapipe}; its 3D counterpart $P_{mesh}$ on $S_{raw}$ is 
acquired by back-projecting $P_{img}$ via the camera parameters of $I_{key}$. Additionally, we leverage a face 
segment~\cite{Yu2021_BiSeNet} to segment out the hair, back-projecting to $S_{raw}$ and removing the hair vertices 
and triangles. Then we use $P_{mesh}$ to roughly estimate the facial plane, calculate its normal direction (nose pointing 
direction) and rotate $S_{raw}$ to align this direction with the z-axis. An axis-aligned bounding box of the face region is then 
computed based on post-rotation $P_{mesh}$ which is applied to acquire $S_{crop}$ (\fg{pipeline_fitting}d). We also segment out the eyes and 
mouth for use in blending with the static texture in Sec. \ref{sec:texgen}, applying back-projection to find the appropriate 
texture-space coordinates.

Once we have $S_{crop}$, $P_{mesh}$ is used to determine the initial rough rigid transformation 
registering $S_{crop}$ to our head model. The FLAME model $F(\theta, t, \beta, \alpha)$ used in
this work has both bones (parameterized by rotation vectors $\theta = \{\theta_g, \theta_1, \dots, \theta_4\}, \theta_i \in \mathbb{R}^3$)
and blendshapes (shape/identity $\beta \in \mathbb{R}^{N_{id}}$ and expression $\alpha \in \mathbb{R}^{N_{expr}}$); for 
this phase we freeze all of the bones and expressions blendshapes, solving for the global rotation $\theta_g$, translation $t \in \mathbb{R}^3$ and shape
parameters. We find these parameters in two steps:
\begin{subequations}\label{equ:fitting}
\begin{align}
    \min_{\theta_g, t} \quad E_{p2p}(S_{lm} - F_{lm} (\theta, t, \beta, \alpha)) \label{fitting_rigid} \\
    \min_{\beta} \quad E_{p2s}(S - F(\theta, t, \beta, \alpha)) \label{fitting_nonrigid}
\end{align}
\end{subequations}
Here $E_{p2p}$ is an $L_2$ point-to-point loss, computed only between vertices 
corresponding to the previously computed landmarks on $S_{crop}$ (denoted $S_{lm}$) and
the corresponding vertices on the FLAME mesh $F_{lm}$. Afterwards, non-rigid fitting is 
done using a point-to-surface loss $E_{p2s}$. For every vertex on the FLAME mesh, we find 
the nearest triangle on $S_{crop}$, and add an $L_2$ loss. The final result is a fitted 
FLAME model (\fg{pipeline_fitting}e) and the corresponding parameters $\theta_g, t, \beta$.

\subsection{Identity-fixed Mesh Tracking}
\label{sec:tracking}
%\textbf{Identity-fixed Mesh Tracking.} 
Now that we have found the identity parameters $\beta$ from Sec. \ref{sec:recons}, we will use
them to prepare the training data for our dynamic texture rendering network. The training video Clip-B consists 
of the subject making a few facial expressions followed by saying a few short sentences. Inspired by the tracker used in~\cite{zielonka2022towards}, while keeping $\beta$ 
fixed, we first adapt the tracking process by keeping camera pose consistently at the world system origin, and in turn to estimate the subject's transformation. We use the camera parameters from $I_{key}$ in Clip-A as initialization to get a global camera calibration $KRT_{global}$ for Clip-B. Clip-B is then split into several chunks to facilitate parallel tracking. Each chunk starts with $KRT_{global}$ as the camera parameters and sequentially estimates the parameters of FLAME expressions, subject's pose, auxiliary lighting, and FLAME texture jointly, as well as further optimizes the camera intrinsics for every frame. The tracking optimization for each frame is accelerated by cautiously reducing the image resolution and texture optimization steps, balancing efficiency and performance.

\begin{figure*}[ht]
\def\svgwidth{\hsize}\import{figs/pipeline_rendering/}{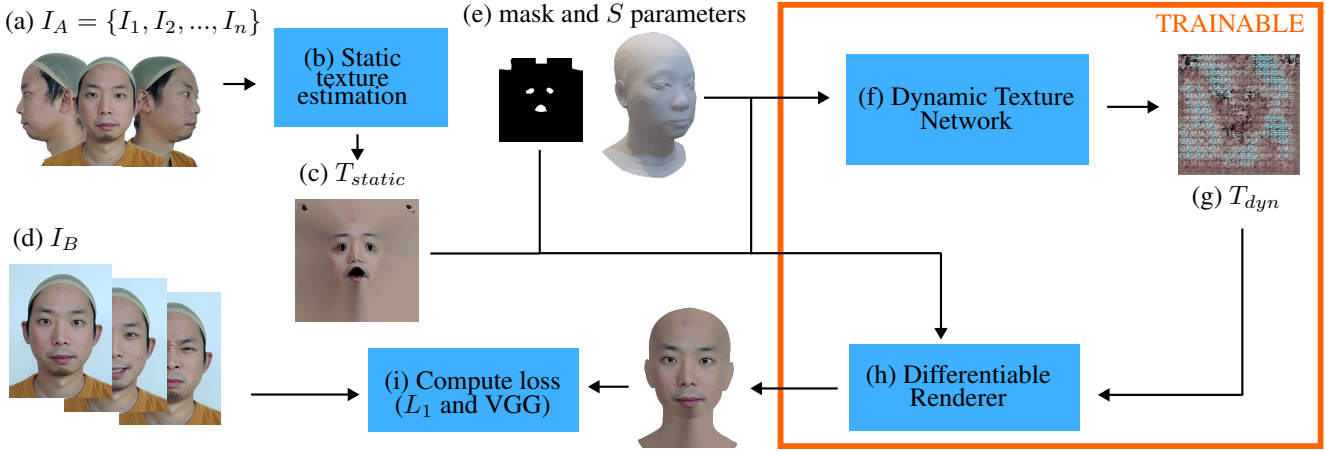}\caption{
The video images from Clip-A (a) are used by the static texture estimation module (b) to 
generate a de-lit and outpainted static texture~(c). Video images (d), and mesh parameters (e) from Clip-B are 
used to train a dynamic texture network (f) which produces a key-region dynamic texture (g) based 
on the head model parameters. The static and dynamic textures are combined with the blending mask, 
and along with the head model are fed into a differentiable renderer (h), the output of which can 
be regressed against the original images~(i).
}\label{fig:pipeline_rendering}
\vspace{-0.10in}
\end{figure*}

\subsection{Texture Generation}
\label{sec:texgen}
%will integrate to the text ... % Fig. TODO shows an overview of our texture generation pipeline. 
The overall texture generation flow starts by recovering a static texture of the subject from the input 
sequence~\fg{pipeline_rendering}a, followed by a de-lighting optimization and outpainting process. This produces a detailed static
texture~\fg{pipeline_rendering}c. The model expression parameters $\alpha$ along with the view angle 
are fed through a dynamic texture rendering network~\fg{pipeline_rendering}f, which produces only the regions
determined by the blending mask acquired in Sec. \ref{sec:recons}. Blending the dynamic texture and the static 
texture, we leverage a differentiable rasterizer $\phi$, PyTorch3D \cite{ravi2020pytorch3d} (\fg{pipeline_rendering}h), 
to render the final textured mesh and back-propagate directly from images.

\subsubsection{Static Texture Generation}
Given the reconstructed mesh $S$~(\fg{pipeline_rendering}b), a set of calibrated images $\textbf{I}_A = \{I_1, I_2, ..., I_n\}$~(\fg{pipeline_rendering}a) acquired through mesh reconstruction (Sec.~\ref{sec:recons}) and the UV texture mapping of $S$, each pixel of the region of interest---face portion in $\textbf{I}_A$ can be projected onto UV space correspondingly to get a UV texture map $T_i$. The unwrapped texture $T_{unwrapped}$ is acquired by $T_{unwrapped}=\frac{1}{n}\sum_{i=1}^{n}T_i$.

Most natural photo sequences will be shot in an environment with non-uniform lighting. This lighting 
will get ``baked'' into $T_{unwrapped}$, which is undesired, as we intend the final texture $T_{static}$ to be used 
from a 3D engine, under potentially different lighting conditions. We therefore approximate the lighting on the photos and remove the approximated light from $T_{unwrapped}$.

To model the light we use spherical harmonics \cite{sloan2023precomputed}.
We perform joint $L_1$ optimization of the first nine coefficients, and the $T_{unwrapped}$’s pixel values.
We add a $\max(\mathit{light}, 1)$ per-pixel regularization term to penalize light becoming too bright and causing clipping for output images.
The light parameters are initialized to a uniform ambient light and the first 50 backpropagation steps only update the light parameters.
Afterward, we start updating the texture pixel values too, which gradually de-lights $T_{unwrapped}$.

To generate the missing skin texture for the neck and rest of the head, we employ patch-based texture synthesis~\cite{Jamriska19,Fiser17} guided by the face portion as an example to synthesize the neck and head textures.
The final output from this step is $T_{static}$ in \fg{pipeline_rendering}c.

\subsubsection{Dynamic Texture Network}

%%%%% Moving to supp
% \begin{figure}
%     \centering
%     \includegraphics[width=0.4\textwidth]{figs/static_dyn_tex/static_dynamic.pdf}
%     \caption{Static texture lacks inner mouth texture and has slightly blurred eyelid texture. The dynamic texture corrects these issues.}
%     \label{fig:sta_dyn}
% \end{figure}
%%%%% Moving to supp

The static texture $T_{static}$ produced above is ready to use in 3D engines when neither close-up look
nor fine-detailed dynamic appearance is required. However, in many cases there are artifacts
around regions crucial for recognizability, such as the eyes and mouth, that can become apparent under
certain views and in motion. The inner part of the mouth can be particularly 
problematic during speech sequences, with the teeth, tongue, and oral cavity all being displayed at different 
times. Our proposed dynamic texture is effective in correcting the mentioned artifacts 
(see the appendix for further discussion and evaluation of static and dynamic texture). 
%Fig.~\ref{fig:sta_dyn} shows the teeth artifacts in the static texture for this subject, 
%and how our dynamic texture can correct them.

%The photorealistic likeliness is already outperform the 3D avatar produced by professional 3D modular artists (a figure?). However, in some key facial regions such as eyes and mouth, finer details are expected since rendering artifacts raised from texture unwrapping are quite noticeable in a dynamic talking and expression sequence. We consequently first propose to realize a dynamic texture specifically targeting at such small portion but crucial regions. 

%\subsubsection{Key-region Dynamic Texture Network}
%Our key idea is to use the the good detail level of the static texture and leverage a small network to add dynamism to the model.
%To do this, we introduce a simple convolutional decoder that generates a texture image from the head model expression parameters $\alpha$.
%We then apply a blending mask  to ensure only the regions of interest are generated; finally combine the generated texture with the 
%static texture using this mask for our final texture. Leveraging a differentiable rasterizer such as , 
%we can apply and render this blended texture during training which prevents seams and discontinuities from occurring due to the blending.

\begin{figure}[ht]
\def\svgwidth{\hsize}\import{figs/rendering_network_arch/}{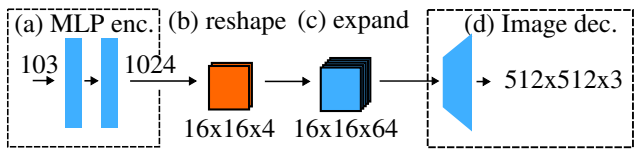}\caption{
The expressions parameters and view angles are fed into the MLP encoder (a) and reshaped (b) for compatibility with convolutional decoders. The number of channels is expanded with a convolutional layer (c) for more network capacity, and is then decoded into an RGB image (d).
}\label{fig:rendering_network_arch}
\vspace{-0.10in}
\end{figure}

\textbf{Network architecture.}
Our network takes as input a vector of the expression and head orientation parameters ($x \in \mathbb{R}^{103}$ for the FLAME model)
and outputs an image the same size as the static texture. See the diagram in~\fg{rendering_network_arch}. The input vector $x$
is fed into a small MLP~(\fg{rendering_network_arch}a) whose output is reshaped to create tensors analogous to small ``images''
with $4$ channels~(\fg{rendering_network_arch}b). These channels are expanded to $64$~(\fg{rendering_network_arch}c) to provide the decoder with more learnable 
parameters with minimal increases to the model size. Finally, these ``images'' are passed through an 
image decoder~(\fg{rendering_network_arch}d) consisting of transposed convolutions to generate the final RGB texture $T_{dyn}$ of the appropriate size. An upsampling 
operation can be added to match the texture resolution. An $L_1$ and $VGG$ loss are computed on the final rendered image:
\begin{align}
    \min_w \sum_i |I_i - \phi(F, T_{dyn}^i(w))| + VGG(I_i, \phi(F, T_{dyn}^i(w))
\end{align}
Where $w$ are the network weights used to generated $T_{dyn}$; $\phi$ and $F$ are the differentiable renderer and the FLAME model,
respectively.

\section{Results}
\label{sec:results}

\subsection{Fitting Comparison}

%%%%% TABLE
\begin{table}[]
\centering
\begin{tabular}{lccccc}
\hline
ID                 & Metric & NextFace & NHA    & INSTA  & Ours   \\ \hline
\multirow{3}{*}{1} & PSNR   & 31.003   & 34.821 & 35.418 & 32.650 \\
                   & SSIM   & 0.947    & 0.963  & 0.964  & 0.968  \\
                   & LPIPS  & 0.063    & 0.015  & 0.018  & 0.046  \\ \hline
\multirow{3}{*}{2} & PSNR   & 30.242   & 35.868 & 38.815 & 32.470 \\
                   & SSIM   & 0.959    & 0.972  & 0.982  & 0.969  \\
                   & LPIPS  & 0.054    & 0.011  & 0.015  & 0.032  \\ \hline
\multirow{3}{*}{3} & PSNR   & 34.283   & 35.594 & 39.137 & 35.240 \\
                   & SSIM   & 0.956    & 0.968  & 0.977  & 0.963  \\
                   & LPIPS  & 0.050    & 0.013  & 0.014  & 0.025  \\ \hline
\end{tabular}
\caption{Quantitative evaluation of the avatar reconstruction on three identities. Numbers are reported as the average of the frames from corresponding test splits.}
\label{tab:render_compare}
\vspace{-0.15in}
\end{table}
%%%%% TABLE

\begin{figure*}[ht]
    \centering
    \includegraphics[width=0.82\textwidth]{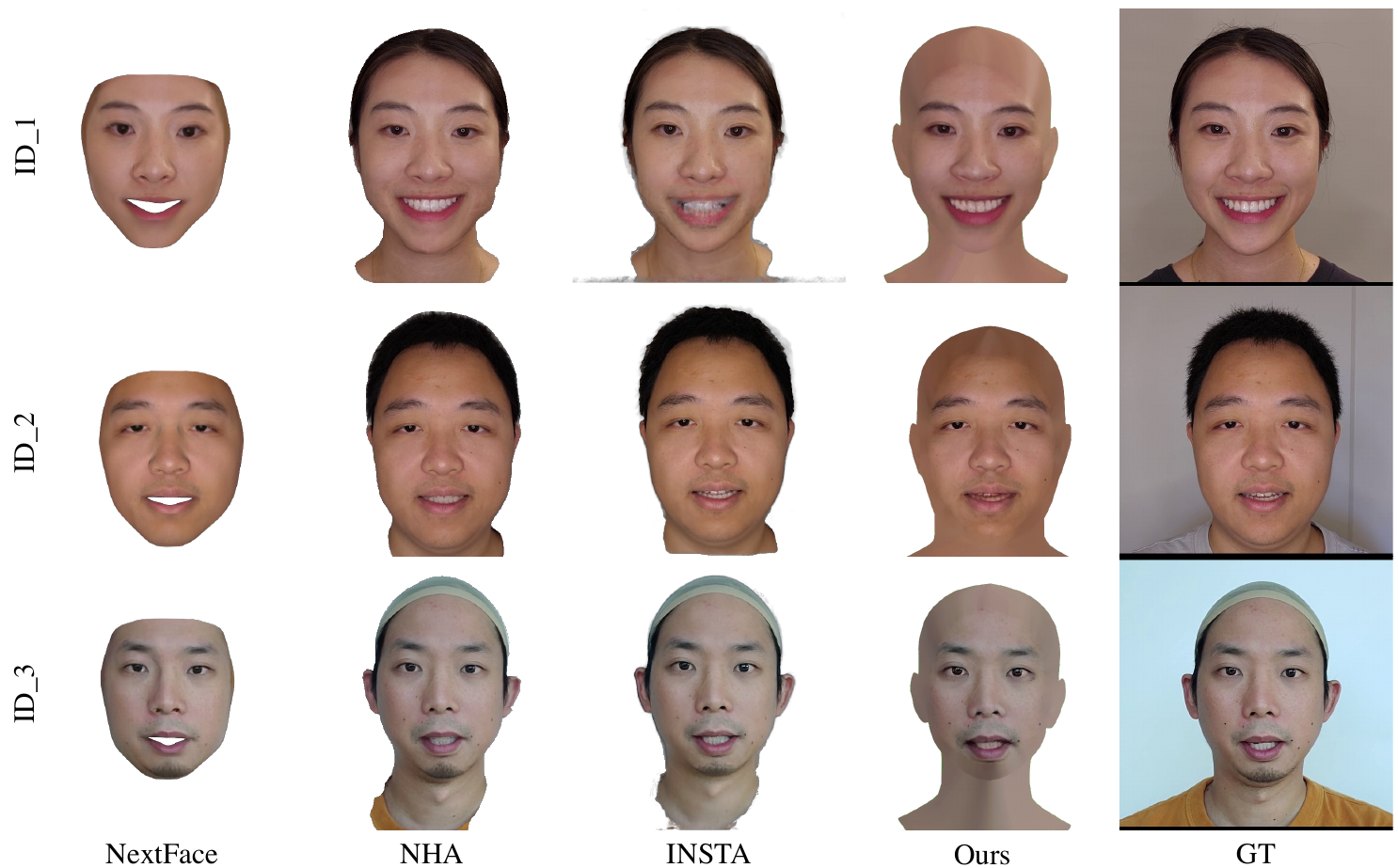}
    \caption{Qualitative comparison of our technique with three state-of-the-art methods: NextFace~\cite{dib2021practical}, Neural Head Avatar (NHA)~\cite{Grassal22_neural_head_avatars}, and Instant Volumetric Head Avatars (INSTA)~\cite{zielonka2023instant}. See the supplementary video for more results.}
    \vspace{-0.10in}
    \label{fig:cross_compare}
\end{figure*}

We compare our fitting result---adapted NDS---with several widely-used off-the-shelf methods: 
DECA~\cite{feng2021learning} and MICA~\cite{zielonka2022towards}. DECA provides per-frame FLAME ID
and expression estimation; thus the ID varies across video frames and exhibits significant 
entanglement between identity and expression, while MICA focuses only on identity, also 
based on the FLAME model. To compare the fitting, we compute the Hausdorff distance between 
the FLAME meshes and the NDS cropped mesh (\fg{pipeline_fitting}d). 
%The errors are colorized in Fig.~\ref{fig:recon_compare}. 
See the supplementary pdf for the quality evaluation of the discussed methods.
Since our reconstruction fits directly to this geometry, our result
yields a lower fitting error. MICA provides a worse fitting compared to our method, and DECA
not only results in a very regularized mesh, but it also outputs erroneous expression and 
bone parameters, leading to more significant errors.

% While Metashape generates extra-high resolution photogrammetry reconstruction, the mesh surface is 
% substantially noisier and since we need to fit FLAME model to the reconstruction, having extra high frequency details does not bring any benefit.

\begin{figure*}[ht]
\def\svgwidth{\hsize}%% Creator: Inkscape 1.3.2 (091e20e, 2023-11-25, custom), www.inkscape.org
%% Accompanies image file 'result_big.pdf' (pdf, eps, ps)
\begingroup%
  \makeatletter%
  \providecommand\color[2][]{%
    \errmessage{(Inkscape) Color is used for the text in Inkscape, but the package 'color.sty' is not loaded}%
    \renewcommand\color[2][]{}%
  }%
  \providecommand\transparent[1]{%
    \errmessage{(Inkscape) Transparency is used (non-zero) for the text in Inkscape, but the package 'transparent.sty' is not loaded}%
    \renewcommand\transparent[1]{}%
  }%
  \providecommand\rotatebox[2]{#2}%
  \newcommand*\fsize{\dimexpr\f@size pt\relax}%
  \newcommand*\lineheight[1]{\fontsize{\fsize}{#1\fsize}\selectfont}%
  \ifx\svgwidth\undefined%
    \setlength{\unitlength}{3778.5bp}%
    \ifx\svgscale\undefined%
      \relax%
    \else%
      \setlength{\unitlength}{\unitlength * \real{\svgscale}}%
    \fi%
  \else%
    \setlength{\unitlength}{\svgwidth}%
  \fi%
  \global\let\svgwidth\undefined%
  \global\let\svgscale\undefined%
  \makeatother%
  \begin{picture}(1,0.38030965)%
    \lineheight{1}%
    \setlength\tabcolsep{0pt}%
    \put(0,0){\includegraphics[width=\unitlength,page=1]{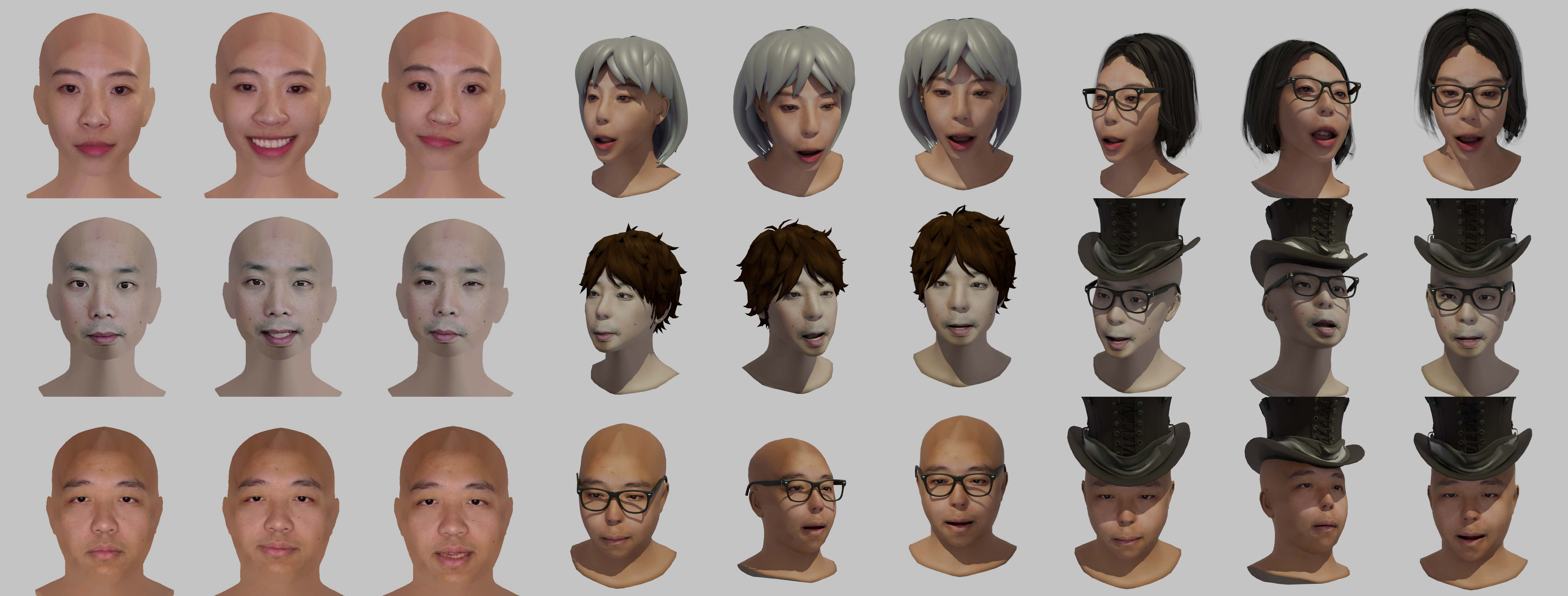}}%
    \put(0.01336096,0.23296695){\color[rgb]{0,0,0}\makebox(0,0)[t]{\lineheight{0}\smash{\begin{tabular}[t]{c}(b)\end{tabular}}}}%
    \put(0.01336096,0.10632939){\color[rgb]{0,0,0}\makebox(0,0)[t]{\lineheight{0}\smash{\begin{tabular}[t]{c}(c)\end{tabular}}}}%
    \put(0.01336096,0.35960451){\color[rgb]{0,0,0}\makebox(0,0)[t]{\lineheight{0}\smash{\begin{tabular}[t]{c}(a)\end{tabular}}}}%
  \end{picture}%
\endgroup%
\caption{
Qualitative results of our 3D avatar generations on three identities. From left to right: (the first 3 columns): rendering results via the PyTorch3D rasterizer with viewing perspective matching the input testing frames; (the last 6 columns): novel-view rendering results with customized facial accessories via the Cycles path tracing engine in Blender. See the supplementary video for more results.
}\label{fig:result_big}
\vspace{-0.15in}
\label{fig:big_result}
\end{figure*}

\subsection{Rendering Comparison}
We compare our final rendering output with several relevant state-of-the-art 3D avatar generation techniques and show that we achieve comparable visual quality in a substantially more practical setting.
Specifically, we compared our rendering result on three identities with three methods: Neural Head Avatar (NHA)~\cite{Grassal22_neural_head_avatars}, Instant Volumetric Head Avatars (INSTA)~\cite{zielonka2023instant} and NextFace~\cite{dib2021practical}. Unlike our method which only reconstructs a face-cropped region (as well as NextFace), both NHA and INSTA aim to reconstruct full head 3D avatar, so we restricted all numeric metric computations to be within the face region. All experiments share the same training/testing split, and are done under their corresponding recommended settings. Tab.~\ref{tab:render_compare} shows the quantitative comparison among these methods.

NHA not only learns view- and expression-dependent neural textures, but also the explicit underlying mesh shape. Though it can yield decent visual quality and better numeric evaluation, its sequential training paradigm with shape deformation, texture learning and joint optimization is quite memory and time consuming (\mytexttilde10 hrs). The resulting mesh reconstruction usually suffers from non-manifold issues, and such topology deficiency will in turn deteriorate the neural texture rendered on it, as shown on ID\_1's cheek contour and ID\_3's ear contour in~Fig.~\ref{fig:cross_compare}. Moreover, the generation is sometimes inaccurate to the utterance as shown on ID\_2.

Different from NHA, which provides an explicit avatar representation, INSTA creates an implicit NeRF-based 3D head avatar which is built upon Instant-NGP~\cite{muller2022instant}. The resulting reconstructions achieve good quantitative performance across all numeric metrics and generally have vivid details, but they also contain slight impurity artifacts as shown on ID\_1 and ID\_2 around neck and hair region respectively in~Fig.~\ref{fig:cross_compare}. Benefiting from the multiresolution hash encoding, INSTA achieves fast avatar generation (\mytexttilde5~min). However, it sometimes has issues dealing with extreme and rare expressions (as shown in the mouth region in ID\_1). It is also non-trivial to plug such NeRF-based implicit avatars into commonly used 3D engines and integrate them with numerous 3D assets. 

NextFace solves a coarse-to-fine optimization formulation to recover a 3D face with 3DMM along with advanced lighting estimation from a single image. It is not designed to take video as input and needs a couple of minutes (\mytexttilde3~min) for reconstructing a single image. We consequently subsample our testing video clips to facilitate the comparison. As shown in~Fig.~\ref{fig:cross_compare}, NextFace tends to generate an over-smoothed facial region that lacks high-frequency details (e.g. wrinkles, freckles) and the inner mouth cavity, resulting the lowest numerical performance.


\subsection{Rendering in a 3D Engine}

As noted in Section \ref{sec:method} of our pipeline description, our system is primarily designed with application-driven objectives rather than purely technique-driven aims. While Neural Radiance Fields (NeRF) rendering continues to gain traction within the academic community due to its promising exploratory and rapid evolutionary progress, integrating such technology into mainstream 3D game engines remains a formidable challenge. Also, there is still little in terms of a robust visual effects (VFX) ecosystem that can effectively harness and elevate NeRF applications within commercial environments.

In contrast, our method adopts a technically conservative approach, yet it offers a commercially viable solution for creating 3D avatars across a broad spectrum of applications. It ensures seamless integration with conventional 3D media production pipelines, enhancing user experience for 3D modelers and designers. This compatibility is vividly demonstrated in Fig.~\ref{fig:big_result}, wherein the right six columns illustrate the diverse rendering effects achievable through integrating our avatars into a 3D engine, highlighting the flexibility of our avatar representation.

\begin{figure}
    \centering
    \includegraphics[width=0.5\textwidth]{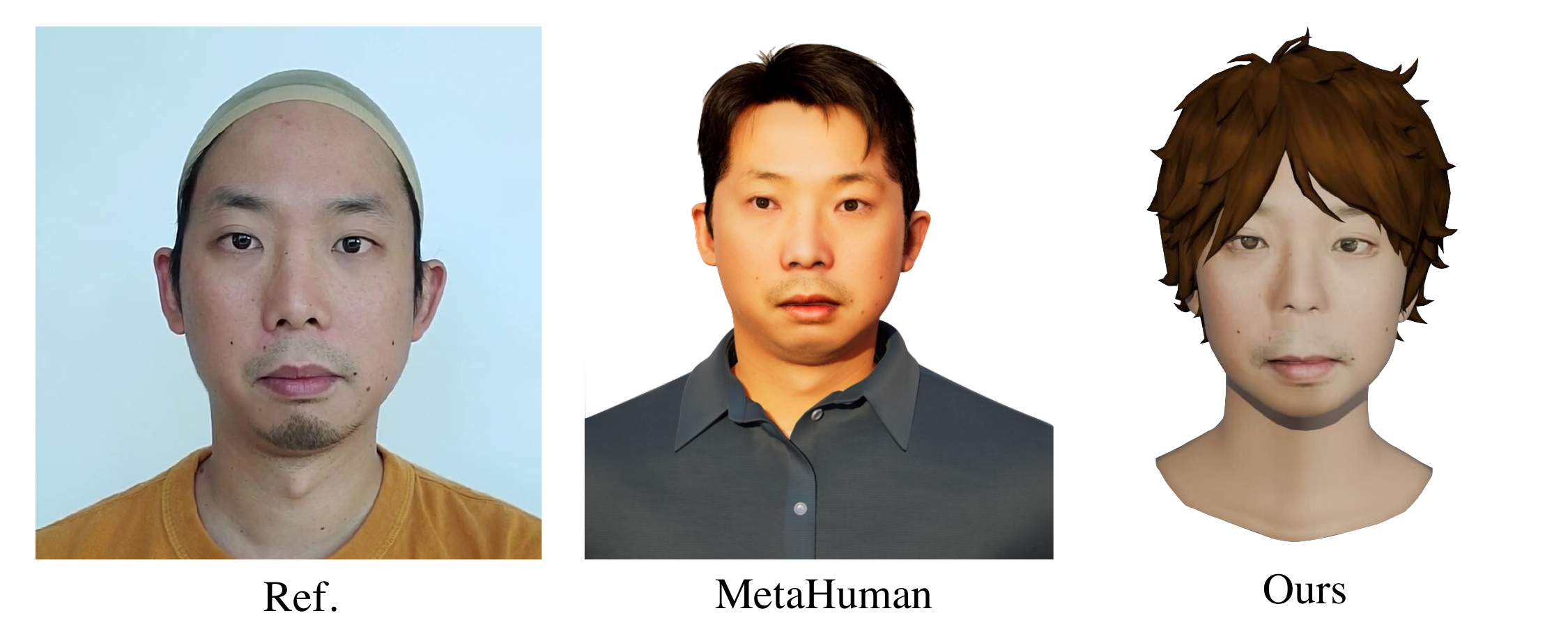}
    \caption{MetaHuman reconstruction vs. ours, both are rendered in a 3D engine.}
    \vspace{-0.2in}
    \label{fig:mh_compare}
\end{figure}

Our learned neural texture significantly enhances visual realism, mitigating the uncanny effects often associated with manually crafted textures. This advancement is showcased in Fig.~\ref{fig:mh_compare}, where a comparison is drawn between our automated pipeline and the MetaHuman creator.
%\footnote{https://www.unrealengine.com/en-US/metahuman}. 
The latter typically requires 2--3 hours of manual effort by a professional 3D modeler collaborating in our study.

Moreover, our method focuses mainly on the immutable features of the facial region, allowing users to customize their avatars with various accessories. This aspect is crucial for VR/AR applications and 3D gaming, where personalization and adaptability are key to user engagement and immersion.

\section{Limitations and Future Work}
\label{sec:limits}
Although our approach presents a novel system that allows for high fidelity photorealistic 3D avatars, there are still certain limitations that present further research opportunity. 
%
%First, 
Our method currently focuses primarily on the facial region using 3DMM, leaving the precise fitting of other parts such as the ears and neck underexplored; that might affect the overall continuity and realism of the avatar, particularly when observed from angles where these areas are prominent.
%
%Second, 
Also, the framework needs to be extended in order to integrate the head 3DMM with a rigged body model within a 3D engine. Such integration is crucial for applications requiring full-body avatars, e.g. virtual reality or fully immersive video gaming, and remains an area for future development.
%
%Third, the inner mouth cavity and texture representation may appear unrealistic at extreme side view angles due to the lack of real underlying topology for teeth and tongue. This limitation can detract from the immersive experience, especially in applications where facial expressions are critical, such as in detailed character animations or virtual interpersonal communications.
%
%Fourth, while our dynamic texture generation approach significantly enhances realism, the outpainting of textures could be improved to appear more natural, especially under varying lighting conditions and complex facial movements; currently, certain parts of the outpainted portion of the texture might appear too smooth lacking the skin pattern, or there might be sharp color changes breaking the overall continuity of the texture.
%Currently, the textures, although advanced, sometimes exhibit areas where improvements in smoothness and continuity could be beneficial.
%

%In future work, we aim to address the above limitations and push the state-of-the-art boundaries of real-time photorealistic avatar generation even further.

\section{Conclusion}
\label{sec:conclude}
In this work, we introduced a streamlined pipeline for creating photorealistic 3D avatars tailored for 3D engines. By leveraging dynamic textures and 3D morphable models, our method reduces the complexity typically associated with high-quality avatar generation. The key advancements of this research include a minimalistic capture process, refined mesh reconstruction from facial data, and efficient texture rendering, all of which significantly enhance the practical deployment of digital avatars in 3D environments. This initiative not only aligns with current trends in digital representation but also paves the way for further developments in immersive digital experiences.

%%%%%%%%% ACKNOWLEDGMENTS
% \ifcamera
% \section*{Acknowledgments}
% We would like to thank ...
% and rest of the team at \href{https://neonlife.ai}{NEON} for their extensive
% engineering, hardware, and design support.
% \fi

%%%%%%%%% Appendix %%%%%%%%%%%%%%%%%%%%%%%%%%%%%%%%%%%%%%%%%%%%%%%%%

%\clearpage
\appendix

%%%%%%%%% COPY PASTE from supp.tex, DO NOT EDIT HERE %%%%%%%%%%%%%%%

%%%% Moving to supp
\section{Capture Setup}\label{sec:cap_set}
The video sequence contains two parts: Clip-A focuses on acquiring the identity information
and Clip-B focuses on extracting the expression and head pose information.
For Clip-A, we found that a sweep of the semi-circle in front of the face was sufficient
for our purposes. There are several ways to capture such a video without another person present.
In this work, we use a simple tripod and rotating chair to ``fake''
the multiview capture by having the subject turn themselves on the
chair without moving their upper body. Users can also take selfie
video clip by holding and sweeping a phone in front of their face,
although we found this requires a little bit of practice to keep their
face in the center. This could be remedied by using a phone gimbal,
but we found the tripod and chair setup to be the most convenient.
Clip-B is a more traditional capture sequence; the camera is fixed and 
we require the subject to make a few facial expressions and read a 
few short sentences. Clip-A is around 20~s and Clip-B is under a minute.
%%%% Moving to supp

%%%% Moving to supp
\section{Implementation Details \& Hardware Requirements}
Our system is implemented using PyTorch and all experiments are tested on NVIDIA GeForce RTX 2080. The mesh reconstruction process, which forms the immutable identity of our avatar creation, completes in less than 5 minutes. For mesh tracking, sequential tracking and parallel tracking (see details in Sec.3.2) requires 15 minutes and 5 minutes, respectively. The size of our compact dynamic texture network is 2~MB. This small footprint allows for rapid deployment and execution, even on less capable hardware. We use Adam optimizer with learning rate $4e^{-4}$. Training the network to achieve satisfactory results takes approximately half an hour, which is roughly $2500$ iterations. Inference is very fast -- taking only about $6.03$ms per frame.

Overall our system is designed to be accessible. It can operate effectively on a standard workstation without the need for high-end computational resources. This accessibility ensures that our method can be used by researchers and developers who may not have access to cutting-edge hardware, democratizing the process of high-quality avatar generation.
%%%% Moving to supp
\section{Fitting Comparison}
%%%% Moving to supp
\begin{figure}[t]
\def\svgwidth{\hsize}%% Creator: Inkscape 1.3.2 (091e20ef0f, 2023-11-25), www.inkscape.org
%% Accompanies image file 'mesh_fitting_figure.pdf' (pdf, eps, ps)
\begingroup%
  \makeatletter%
  \providecommand\color[2][]{%
    \errmessage{(Inkscape) Color is used for the text in Inkscape, but the package 'color.sty' is not loaded}%
    \renewcommand\color[2][]{}%
  }%
  \providecommand\transparent[1]{%
    \errmessage{(Inkscape) Transparency is used (non-zero) for the text in Inkscape, but the package 'transparent.sty' is not loaded}%
    \renewcommand\transparent[1]{}%
  }%
  \providecommand\rotatebox[2]{#2}%
  \newcommand*\fsize{\dimexpr\f@size pt\relax}%
  \newcommand*\lineheight[1]{\fontsize{\fsize}{#1\fsize}\selectfont}%
  \ifx\svgwidth\undefined%
    \setlength{\unitlength}{237.59999657bp}%
    \ifx\svgscale\undefined%
      \relax%
    \else%
      \setlength{\unitlength}{\unitlength * \real{\svgscale}}%
    \fi%
  \else%
    \setlength{\unitlength}{\svgwidth}%
  \fi%
  \global\let\svgwidth\undefined%
  \global\let\svgscale\undefined%
  \makeatother%
  \begin{picture}(1,0.75757577)%
    \lineheight{1}%
    \setlength\tabcolsep{0pt}%
    \put(0,0){\includegraphics[width=\unitlength,page=1]{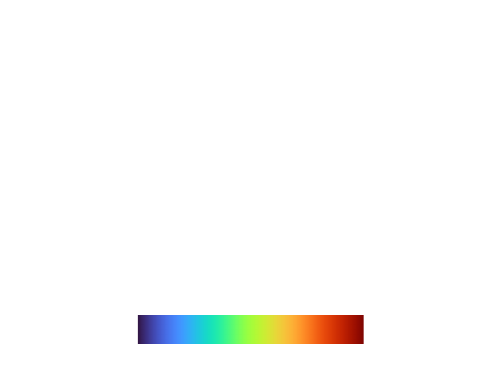}}%
    \put(0.45130139,0.16887464){\color[rgb]{0,0,0}\makebox(0,0)[lt]{\lineheight{1.25}\smash{\begin{tabular}[t]{l}MICA\\\end{tabular}}}}%
    \put(0.77385089,0.16863503){\color[rgb]{0,0,0}\makebox(0,0)[lt]{\lineheight{1.25}\smash{\begin{tabular}[t]{l}Our fitting\end{tabular}}}}%
    \put(0.24816258,0.03184687){\color[rgb]{0,0,0}\makebox(0,0)[lt]{\lineheight{1.25}\smash{\begin{tabular}[t]{l}0 mm\end{tabular}}}}%
    \put(0.6774181,0.03231772){\color[rgb]{0,0,0}\makebox(0,0)[lt]{\lineheight{1.25}\smash{\begin{tabular}[t]{l}1 mm\end{tabular}}}}%
    \put(0,0){\includegraphics[width=\unitlength,page=2]{mesh_fitting_figure_x.pdf}}%
    \put(0.09000093,0.1688786){\color[rgb]{0,0,0}\makebox(0,0)[lt]{\lineheight{1.25}\smash{\begin{tabular}[t]{l}DECA\\\end{tabular}}}}%
  \end{picture}%
\endgroup%
\caption{
Qualitative visualization of static identity mesh fitting among three methods.
}\label{fig:mesh_fitting_figure}
\label{fig:recon_compare}
\end{figure}
%%%% Moving to supp
Fig.~\ref{fig:recon_compare} shows Hausdorff distance quality mapping comparison among three methods mentioned in Sec.4.3.1. The color bar shows the error scale. The error is the Hausdorff distance computed against $S_{crop}$ described in Sec.3.1.

\section{Static Texture vs Dynamic Texture}
%%%%% Moving to supp
\begin{figure}
    \centering
    \includegraphics[width=0.4\textwidth]{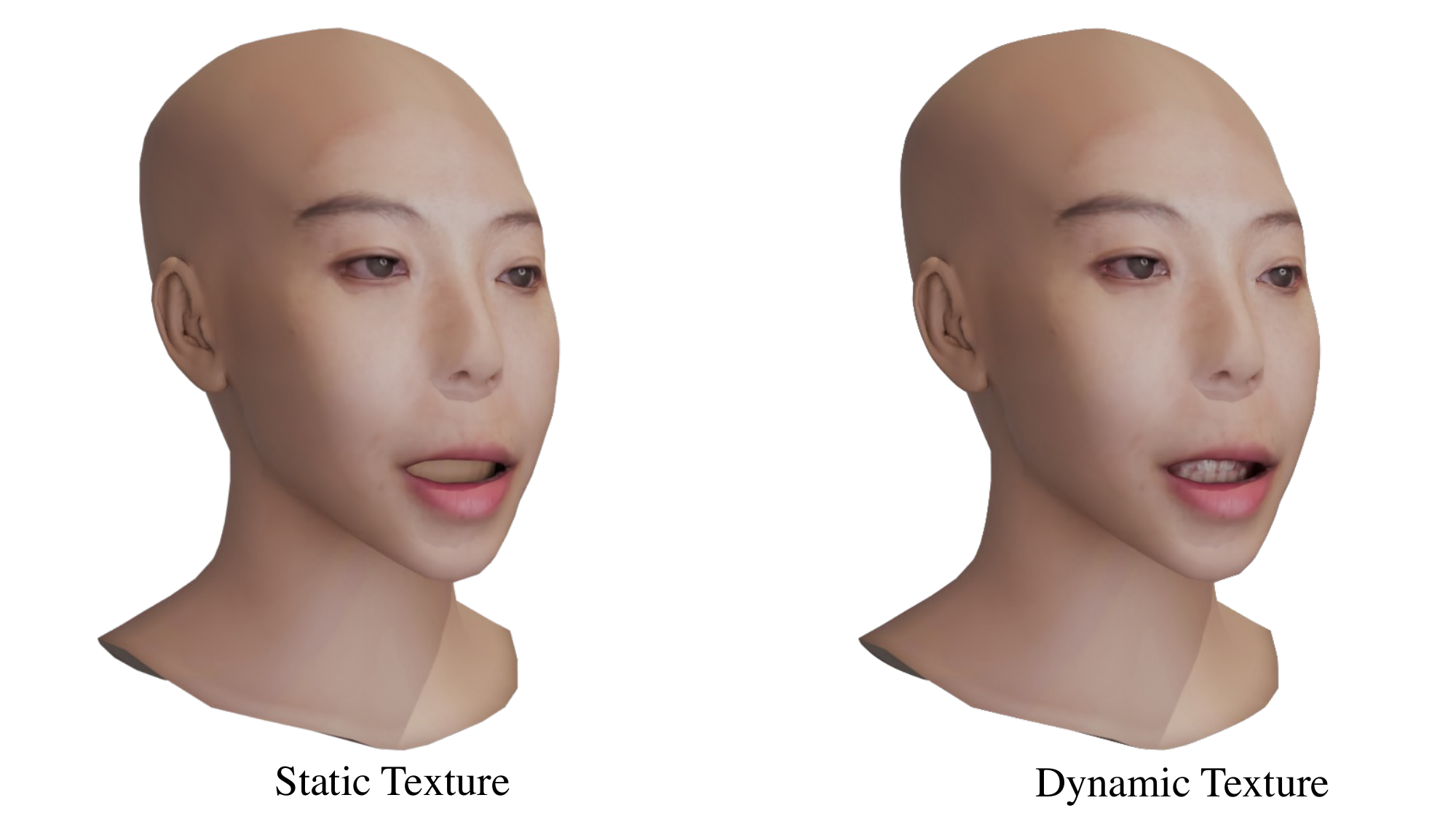}
    \caption{Static texture lacks inner mouth texture and has slightly blurred eyelid texture. The dynamic texture corrects these issues.}
    \label{fig:sta_dyn}
\end{figure}
%%%%% Moving to supp
%%%% Moving to supp
Fig.~\ref{fig:sta_dyn} shows the qualitative visualization of of static and dynamic texture of one subject. One can see that the tooth and eyelid artifacts in the static texture are corrected by our dynamic texture.

%%%%%%%%% REFERENCES
{\small
\bibliographystyle{ieee_fullname}
\bibliography{main}

\begin{thebibliography}{10}\itemsep=-1pt

\bibitem{Blanz1999AMM}
Volker Blanz and Thomas Vetter.
\newblock A morphable model for the synthesis of 3d faces.
\newblock In {\em Proceedings of the 26th Annual Conference on Computer
  Graphics and Interactive Techniques}, pages 187--194, 1999.

\bibitem{blanz2023morphable}
Volker Blanz and Thomas Vetter.
\newblock A morphable model for the synthesis of 3d faces.
\newblock In {\em Seminal Graphics Papers: Pushing the Boundaries, Volume 2},
  pages 157--164. 2023.

\bibitem{Bregler1997}
Christoph Bregler, Michele Covell, and Malcolm Slaney.
\newblock Video rewrite: Driving visual speech with audio.
\newblock {\em ACM Transactions on Graphics}, 31:353--360, 01 1997.

\bibitem{cao22}
Chen Cao, Tomas Simon, Jin~Kyu Kim, Gabe Schwartz, Michael Zollhoefer,
  Shun-Suke Saito, Stephen Lombardi, Shih-En Wei, Danielle Belko, Shoou-I Yu,
  Yaser Sheikh, and Jason Saragih.
\newblock Authentic volumetric avatars from a phone scan.
\newblock {\em ACM Transactions on Graphics}, 41(4), 2022.

\bibitem{cao2016real}
Chen Cao, Hongzhi Wu, Yanlin Weng, Tianjia Shao, and Kun Zhou.
\newblock Real-time facial animation with image-based dynamic avatars.
\newblock {\em ACM Transactions on Graphics}, 35(4), 2016.

\bibitem{chan2022efficient}
Eric~R Chan, Connor~Z Lin, Matthew~A Chan, Koki Nagano, Boxiao Pan, Shalini
  De~Mello, Orazio Gallo, Leonidas~J Guibas, Jonathan Tremblay, Sameh Khamis,
  et~al.
\newblock Efficient geometry-aware 3d generative adversarial networks.
\newblock In {\em Proceedings of the IEEE/CVF Conference on Computer Vision and
  Pattern Recognition}, pages 16123--16133, 2022.

\bibitem{Chen2019_hierarchical}
Lele Chen, Ross~K. Maddox, Zhiyao Duan, and Chenliang Xu.
\newblock Hierarchical cross-modal talking face generation with dynamic
  pixel-wise loss.
\newblock {\em 2019 IEEE/CVF Conference on Computer Vision and Pattern
  Recognition (CVPR)}, pages 7824--7833, 2019.

\bibitem{Cudeiro2019capture}
Daniel Cudeiro, Timo Bolkart, Cassidy Laidlaw, Anurag Ranjan, and Michael~J
  Black.
\newblock Capture, learning, and synthesis of 3d speaking styles.
\newblock In {\em Proceedings of IEEE Conference on Computer Vision and Pattern
  Recognition}, pages 10101--10111, 2019.

\bibitem{danvevcek2022emoca}
Radek Dan{\v{e}}{\v{c}}ek, Michael~J Black, and Timo Bolkart.
\newblock Emoca: Emotion driven monocular face capture and animation.
\newblock In {\em Proceedings of the IEEE/CVF Conference on Computer Vision and
  Pattern Recognition}, pages 20311--20322, 2022.

\bibitem{dhariwal21}
Prafulla Dhariwal and Alexander~Quinn Nichol.
\newblock Diffusion models beat gans on image synthesis.
\newblock In {\em Advances in Neural Information Processing Systems, NeurIPS},
  pages 8780--8794, 2021.

\bibitem{dib2021practical}
Abdallah Dib, Gaurav Bharaj, Junghyun Ahn, C{\'e}dric Th{\'e}bault, Philippe
  Gosselin, Marco Romeo, and Louis Chevallier.
\newblock Practical face reconstruction via differentiable ray tracing.
\newblock In {\em Computer Graphics Forum}, volume~40, pages 153--164. Wiley
  Online Library, 2021.

\bibitem{dib2021towards}
Abdallah Dib, Cedric Thebault, Junghyun Ahn, Philippe-Henri Gosselin, Christian
  Theobalt, and Louis Chevallier.
\newblock Towards high fidelity monocular face reconstruction with rich
  reflectance using self-supervised learning and ray tracing.
\newblock In {\em Proceedings of the IEEE/CVF International Conference on
  Computer Vision}, pages 12819--12829, 2021.

\bibitem{du23}
Chenpng Du, Qi Chen, Tianyu He, Xuejiao Tan, Xie Chen, K. Yu, Sheng Zhao, and
  Jiang Bian.
\newblock Dae-talker: High fidelity speech-driven talking face generation with
  diffusion autoencoder.
\newblock {\em Proceedings of ACM International Conference on Multimedia},
  2023.

\bibitem{egger20203d}
Bernhard Egger, William~AP Smith, Ayush Tewari, Stefanie Wuhrer, Michael
  Zollhoefer, Thabo Beeler, Florian Bernard, Timo Bolkart, Adam Kortylewski,
  Sami Romdhani, et~al.
\newblock 3d morphable face models—past, present, and future.
\newblock {\em ACM Transactions on Graphics (ToG)}, 39(5):1--38, 2020.

\bibitem{esser20}
Patrick Esser, Robin Rombach, and Bj{\"o}rn Ommer.
\newblock Taming transformers for high-resolution image synthesis.
\newblock {\em IEEE/CVF Conference on Computer Vision and Pattern Recognition
  (CVPR)}, pages 12868--12878, 2020.

\bibitem{feng2021learning}
Yao Feng, Haiwen Feng, Michael~J Black, and Timo Bolkart.
\newblock Learning an animatable detailed 3d face model from in-the-wild
  images.
\newblock {\em ACM Transactions on Graphics (ToG)}, 40(4):1--13, 2021.

\bibitem{filntisis2023spectre}
Panagiotis~P Filntisis, George Retsinas, Foivos Paraperas-Papantoniou,
  Athanasios Katsamanis, Anastasios Roussos, and Petros Maragos.
\newblock Spectre: Visual speech-informed perceptual 3d facial expression
  reconstruction from videos.
\newblock In {\em Proceedings of the IEEE/CVF Conference on Computer Vision and
  Pattern Recognition}, pages 5744--5754, 2023.

\bibitem{Fiser17}
Jakub Fi\v{s}er, Ond\v{r}ej Jamri\v{s}ka, David Simons, Eli Shechtman, Jingwan
  Lu, Paul Asente, Michal Luk\'{a}\v{c}, and Daniel S\'{y}kora.
\newblock Example-based synthesis of stylized facial animations.
\newblock {\em ACM Transactions on Graphics}, 36(4):155, 2017.

\bibitem{gafni2021dynamic}
Guy Gafni, Justus Thies, Michael Zollhofer, and Matthias Nie{\ss}ner.
\newblock Dynamic neural radiance fields for monocular 4d facial avatar
  reconstruction.
\newblock In {\em Proceedings of the IEEE/CVF Conference on Computer Vision and
  Pattern Recognition}, pages 8649--8658, 2021.

\bibitem{giebenhain2023learning}
Simon Giebenhain, Tobias Kirschstein, Markos Georgopoulos, Martin R{\"u}nz,
  Lourdes Agapito, and Matthias Nie{\ss}ner.
\newblock Learning neural parametric head models.
\newblock In {\em Proceedings of the IEEE/CVF Conference on Computer Vision and
  Pattern Recognition}, pages 21003--21012, 2023.

\bibitem{Goodfellow14}
Ian~J. Goodfellow, Jean Pouget-Abadie, Mehdi Mirza, Bing Xu, David
  Warde-Farley, Sherjil Ozair, Aaron~C. Courville, and Yoshua Bengio.
\newblock Generative adversarial nets.
\newblock In {\em Advances in Neural Information Processing Systems}, pages
  2672--2680, 2014.

\bibitem{Grassal22_neural_head_avatars}
Philip-William Grassal, Malte Prinzler, Titus Leistner, Carsten Rother,
  Matthias Nie{\ss}ner, and Justus Thies.
\newblock Neural head avatars from monocular rgb videos.
\newblock In {\em Proceedings of IEEE Conference on Computer Vision and Pattern
  Recognition}, pages 18653--18664, 2022.

\bibitem{gu2021stylenerf}
Jiatao Gu, Lingjie Liu, Peng Wang, and Christian Theobalt.
\newblock Stylenerf: A style-based 3d-aware generator for high-resolution image
  synthesis.
\newblock {\em arXiv preprint arXiv:2110.08985}, 2021.

\bibitem{guo2021ad}
Yudong Guo, Keyu Chen, Sen Liang, Yong-Jin Liu, Hujun Bao, and Juyong Zhang.
\newblock Ad-nerf: Audio driven neural radiance fields for talking head
  synthesis.
\newblock In {\em Proceedings of the IEEE/CVF International Conference on
  Computer Vision}, pages 5784--5794, 2021.

\bibitem{he23}
Tianyu He, Junliang Guo, Runyi Yu, Yuchi Wang, Jialiang Zhu, Kaikai An, Leyi
  Li, Xu Tan, Chunyu Wang, Han Hu, HsiangTao Wu, Sheng Zhao, and Jiang Bian.
\newblock {GAIA}: Zero-shot talking avatar generation.
\newblock {\em arXiv preprint arXiv:2311.15230}, 2023.

\bibitem{ho20}
Jonathan Ho, Ajay Jain, and Pieter Abbeel.
\newblock Denoising diffusion probabilistic models.
\newblock In {\em Advances in Neural Information Processing Systems},
  volume~33, pages 6840--6851, 2020.

\bibitem{hong2022depth}
Fa-Ting Hong, Longhao Zhang, Li Shen, and Dan Xu.
\newblock Depth-aware generative adversarial network for talking head video
  generation.
\newblock In {\em Proceedings of IEEE Conference on Computer Vision and Pattern
  Recognition}, pages 3397--3406, 2022.

\bibitem{hong2022headnerf}
Yang Hong, Bo Peng, Haiyao Xiao, Ligang Liu, and Juyong Zhang.
\newblock Headnerf: A real-time nerf-based parametric head model.
\newblock In {\em Proceedings of the IEEE/CVF Conference on Computer Vision and
  Pattern Recognition}, pages 20374--20384, 2022.

\bibitem{hu2017avatar}
Liwen Hu, Shunsuke Saito, Lingyu Wei, Koki Nagano, Jaewoo Seo, Jens Fursund,
  Iman Sadeghi, Carrie Sun, Yen-Chun Chen, and Hao Li.
\newblock Avatar digitization from a single image for real-time rendering.
\newblock {\em ACM Transactions on Graphics (ToG)}, 36(6):1--14, 2017.

\bibitem{Jamriska19}
Ond\v{r}ej Jamri\v{s}ka, \v{S}\'{a}rka Sochorov\'{a}, Ond\v{r}ej Texler, Michal
  Luk\'{a}\v{c}, Jakub Fi\v{s}er, Jingwan Lu, Eli Shechtman, and Daniel
  S\'{y}kora.
\newblock Stylizing video by example.
\newblock {\em ACM Transactions on Graphics}, 38(4):107, 2019.

\bibitem{Khakhulin2022ROME}
Taras Khakhulin, Vanessa Sklyarova, Victor Lempitsky, and Egor Zakharov.
\newblock Realistic one-shot mesh-based head avatars.
\newblock In {\em Proceedings of European Conference on Computer Vision}, 2022.

\bibitem{Knothe2011}
Reinhard Knothe, Brian Amberg, Sami Romdhani, Volker Blanz, and Thomas Vetter.
\newblock Morphable models of faces.
\newblock In {\em Handbook of Face Recognition}, pages 137--168. Springer
  London, 2011.

\bibitem{lewis2014practice}
John~P Lewis, Ken Anjyo, Taehyun Rhee, Mengjie Zhang, Frederic~H Pighin, and
  Zhigang Deng.
\newblock Practice and theory of blendshape facial models.
\newblock {\em Eurographics (State of the Art Reports)}, 1(8):2, 2014.

\bibitem{li2020learning}
Ruilong Li, Karl Bladin, Yajie Zhao, Chinmay Chinara, Owen Ingraham, Pengda
  Xiang, Xinglei Ren, Pratusha Prasad, Bipin Kishore, Jun Xing, et~al.
\newblock Learning formation of physically-based face attributes.
\newblock In {\em Proceedings of the IEEE/CVF conference on computer vision and
  pattern recognition}, pages 3410--3419, 2020.

\bibitem{Li17_FLAME}
Tianye Li, Timo Bolkart, Michael.~J. Black, Hao Li, and Javier Romero.
\newblock Learning a model of facial shape and expression from {4D} scans.
\newblock {\em ACM Transactions on Graphics}, 36(6), 2017.

\bibitem{li2023generalizable}
Xueting Li, Shalini De~Mello, Sifei Liu, Koki Nagano, Umar Iqbal, and Jan
  Kautz.
\newblock Generalizable one-shot neural head avatar.
\newblock {\em arXiv preprint arXiv:2306.08768}, 2023.

\bibitem{Lombardi18}
Stephen Lombardi, Jason Saragih, Tomas Simon, and Yaser Sheikh.
\newblock Deep appearance models for face rendering.
\newblock {\em ACM Transactions on Graphics}, 37(4):68:1--68:13, 2018.

\bibitem{Lombardi19}
Stephen Lombardi, Tomas Simon, Jason Saragih, Gabriel Schwartz, Andreas
  Lehrmann, and Yaser Sheikh.
\newblock Neural volumes: Learning dynamic renderable volumes from images.
\newblock {\em ACM Transactions on Graphics}, 38(4):65:1--65:14, 2019.

\bibitem{Lombardi21}
Stephen Lombardi, Tomas Simon, Gabriel Schwartz, Michael Zollhoefer, Yaser
  Sheikh, and Jason Saragih.
\newblock Mixture of volumetric primitives for efficient neural rendering.
\newblock {\em ACM Transactions on Graphics}, 40(4), 2021.

\bibitem{lugaresi2019mediapipe}
Camillo Lugaresi, Jiuqiang Tang, Hadon Nash, Chris McClanahan, Esha Uboweja,
  Michael Hays, Fan Zhang, Chuo-Ling Chang, Ming Yong, Juhyun Lee, et~al.
\newblock Mediapipe: A framework for perceiving and processing reality.
\newblock In {\em Third workshop on computer vision for AR/VR at IEEE computer
  vision and pattern recognition (CVPR)}, volume 2019, 2019.

\bibitem{luo2021normalized}
Huiwen Luo, Koki Nagano, Han-Wei Kung, Qingguo Xu, Zejian Wang, Lingyu Wei,
  Liwen Hu, and Hao Li.
\newblock Normalized avatar synthesis using stylegan and perceptual refinement.
\newblock In {\em Proceedings of the IEEE/CVF Conference on Computer Vision and
  Pattern Recognition}, pages 11662--11672, 2021.

\bibitem{ma2023otavatar}
Zhiyuan Ma, Xiangyu Zhu, Guo-Jun Qi, Zhen Lei, and Lei Zhang.
\newblock Otavatar: One-shot talking face avatar with controllable tri-plane
  rendering.
\newblock In {\em Proceedings of the IEEE/CVF Conference on Computer Vision and
  Pattern Recognition}, pages 16901--16910, 2023.

\bibitem{mihajlovic2022keypointnerf}
Marko Mihajlovic, Aayush Bansal, Michael Zollhoefer, Siyu Tang, and Shunsuke
  Saito.
\newblock Keypointnerf: Generalizing image-based volumetric avatars using
  relative spatial encoding of keypoints.
\newblock In {\em European conference on computer vision}, pages 179--197.
  Springer, 2022.

\bibitem{Mildenhall2020nerf}
Ben Mildenhall, Pratul~P. Srinivasan, Matthew Tancik, Jonathan~T. Barron, Ravi
  Ramamoorthi, and Ren Ng.
\newblock Nerf: Representing scenes as neural radiance fields for view
  synthesis.
\newblock In {\em Proceedings of European Conference on Computer Vision}, 2020.

\bibitem{muller2022instant}
Thomas M{\"u}ller, Alex Evans, Christoph Schied, and Alexander Keller.
\newblock Instant neural graphics primitives with a multiresolution hash
  encoding.
\newblock {\em ACM Transactions on Graphics (ToG)}, 41(4):1--15, 2022.

\bibitem{nagano2018pagan}
Koki Nagano, Jaewoo Seo, Jun Xing, Lingyu Wei, Zimo Li, Shunsuke Saito, Aviral
  Agarwal, Jens Fursund, Hao Li, Richard Roberts, et~al.
\newblock pagan: real-time avatars using dynamic textures.
\newblock {\em ACM Trans. Graph.}, 37(6):258, 2018.

\bibitem{nehvi2023360deg}
Jalees Nehvi, Berna Kabadayi, Julien Valentin, and Justus Thies.
\newblock 360$^{\circ}$ volumetric portrait avatar.
\newblock 2023.

\bibitem{nirkin2019fsgan}
Yuval Nirkin, Yosi Keller, and Tal Hassner.
\newblock {FSGAN}: Subject agnostic face swapping and reenactment.
\newblock In {\em Proceedings of IEEE International Conference on Computer
  Vision}, pages 7184--7193, 2019.

\bibitem{nirkin2022fsganv2}
Yuval Nirkin, Yosi Keller, and Tal Hassner.
\newblock {FSGANv2}: Improved subject agnostic face swapping and reenactment.
\newblock {\em IEEE Transactions on Pattern Analysis and Machine Intelligence},
  2022.

\bibitem{park2021nerfies}
Keunhong Park, Utkarsh Sinha, Jonathan~T Barron, Sofien Bouaziz, Dan~B Goldman,
  Steven~M Seitz, and Ricardo Martin-Brualla.
\newblock Nerfies: Deformable neural radiance fields.
\newblock In {\em Proceedings of the IEEE/CVF International Conference on
  Computer Vision}, pages 5865--5874, 2021.

\bibitem{park2021hypernerf}
Keunhong Park, Utkarsh Sinha, Peter Hedman, Jonathan~T Barron, Sofien Bouaziz,
  Dan~B Goldman, Ricardo Martin-Brualla, and Steven~M Seitz.
\newblock Hypernerf: A higher-dimensional representation for topologically
  varying neural radiance fields.
\newblock {\em arXiv preprint arXiv:2106.13228}, 2021.

\bibitem{Prajwal2020lip}
KR Prajwal, Rudrabha Mukhopadhyay, Vinay~P Namboodiri, and CV Jawahar.
\newblock A lip sync expert is all you need for speech to lip generation in the
  wild.
\newblock In {\em Proceedings of the 28th ACM International Conference on
  Multimedia}, pages 484--492, 2020.

\bibitem{ramesh21}
Aditya Ramesh, Mikhail Pavlov, Gabriel Goh, Scott Gray, Chelsea Voss, Alec
  Radford, Mark Chen, and Ilya Sutskever.
\newblock Zero-shot text-to-image generation.
\newblock In {\em Proceedings of the 38th International Conference on Machine
  Learning}, volume 139, pages 8821--8831, 2021.

\bibitem{ravi2020pytorch3d}
Nikhila Ravi, Jeremy Reizenstein, David Novotny, Taylor Gordon, Wan-Yen Lo,
  Justin Johnson, and Georgia Gkioxari.
\newblock Accelerating 3d deep learning with pytorch3d.
\newblock {\em arXiv:2007.08501}, 2020.

\bibitem{Ravichandran23-CVPR}
Siddarth Ravichandran, Ond\v{r}ej Texler, Dimitar Dinev, and Hyun~Jae Hang.
\newblock Synthesizing photorealistic virtual humans through cross-modal
  disentanglement.
\newblock 2023.

\bibitem{Richard_2021_ICCV}
Alexander Richard, Michael Zollh\"ofer, Yandong Wen, Fernando de~la Torre, and
  Yaser Sheikh.
\newblock Meshtalk: 3d face animation from speech using cross-modality
  disentanglement.
\newblock In {\em Proceedings of IEEE International Conference on Computer
  Vision}, pages 1173--1182, October 2021.

\bibitem{rombach21}
Robin Rombach, A. Blattmann, Dominik Lorenz, Patrick Esser, and Bj{\"o}rn
  Ommer.
\newblock High-resolution image synthesis with latent diffusion models.
\newblock {\em IEEE/CVF Conference on Computer Vision and Pattern Recognition},
  pages 10674--10685, 2021.

\bibitem{schoenberger2016sfm}
Johannes~Lutz Sch\"{o}nberger and Jan-Michael Frahm.
\newblock {Structure-from-Motion Revisited}.
\newblock In {\em Conference on Computer Vision and Pattern Recognition
  (CVPR)}, 2016.

\bibitem{shen23}
Shuai Shen, Wenliang Zhao, Zibin Meng, Wanhua Li, Zheng Zhu, Jie Zhou, and
  Jiwen Lu.
\newblock Difftalk: Crafting diffusion models for generalized audio-driven
  portraits animation.
\newblock In {\em Proceedings of IEEE Conference on Computer Vision and Pattern
  Recognition}, 2023.

\bibitem{shi2021lifting}
Yichun Shi, Divyansh Aggarwal, and Anil~K Jain.
\newblock Lifting 2d stylegan for 3d-aware face generation.
\newblock In {\em Proceedings of the IEEE/CVF conference on computer vision and
  pattern recognition}, pages 6258--6266, 2021.

\bibitem{sloan2023precomputed}
Peter-Pike Sloan, Jan Kautz, and John Snyder.
\newblock Precomputed radiance transfer for real-time rendering in dynamic,
  low-frequency lighting environments.
\newblock In {\em Seminal Graphics Papers: Pushing the Boundaries, Volume 2},
  pages 339--348. 2023.

\bibitem{song21}
Jiaming Song, Chenlin Meng, and Stefano Ermon.
\newblock Denoising diffusion implicit models.
\newblock In {\em Proceedings of International Conference on Learning
  Representations}, 2021.

\bibitem{song2018talking}
Yang Song, Jingwen Zhu, Dawei Li, Xiaolong Wang, and Hairong Qi.
\newblock Talking face generation by conditional recurrent adversarial network.
\newblock {\em arXiv preprint arXiv:1804.04786}, 2018.

\bibitem{stypulkowski23}
Micha{\l} Stypu{\l}kowski, Konstantinos Vougioukas, Sen He, Maciej Zieba,
  Stavros Petridis, and Maja Pantic.
\newblock Diffused heads: Diffusion models beat gans on talking-face
  generation.
\newblock In {\em https://arxiv.org/abs/2301.03396}, 2023.

\bibitem{Suwajanakorn2017}
Supasorn Suwajanakorn, Steven~M. Seitz, and Ira Kemelmacher-Shlizerman.
\newblock Synthesizing obama: Learning lip sync from audio.
\newblock {\em ACM Transactions on Graphics}, 36(4), 2017.

\bibitem{Thies2020neural}
Justus Thies, Mohamed Elgharib, Ayush Tewari, Christian Theobalt, and Matthias
  Nie{\ss}ner.
\newblock Neural voice puppetry: Audio-driven facial reenactment.
\newblock In {\em Proceedings of European Conference on Computer Vision}, pages
  716--731, 2020.

\bibitem{Thies2016face2face}
Justus Thies, Michael Zollhofer, Marc Stamminger, Christian Theobalt, and
  Matthias Nie{\ss}ner.
\newblock Face2face: Real-time face capture and reenactment of rgb videos.
\newblock In {\em Proceedings of IEEE Conference on Computer Vision and Pattern
  Recognition}, pages 2387--2395, 2016.

\bibitem{Vougioukas19}
Konstantinos Vougioukas, Stavros Petridis, and Maja Pantic.
\newblock End-to-end speech-driven realistic facial animation with temporal
  gans.
\newblock In {\em Proceedings of IEEE Conference on Computer Vision and Pattern
  Recognition, Workshops}, June 2019.

\bibitem{wang2021facevid2vid}
Ting-Chun Wang, Arun Mallya, and Ming-Yu Liu.
\newblock One-shot free-view neural talking-head synthesis for video
  conferencing.
\newblock In {\em Proceedings of IEEE Conference on Computer Vision and Pattern
  Recognition}, 2021.

\bibitem{worchel2022multi}
Markus Worchel, Rodrigo Diaz, Weiwen Hu, Oliver Schreer, Ingo Feldmann, and
  Peter Eisert.
\newblock Multi-view mesh reconstruction with neural deferred shading.
\newblock In {\em Proceedings of the IEEE/CVF Conference on Computer Vision and
  Pattern Recognition}, pages 6187--6197, 2022.

\bibitem{wayne2018reenactgan}
Wayne Wu, Yunxuan Zhang, Cheng Li, Chen Qian, and Chen~Change Loy.
\newblock Reenactgan: Learning to reenact faces via boundary transfer.
\newblock In {\em Proceedings of European Conference on Computer Vision}, 2018.

\bibitem{xu2023latentavatar}
Yuelang Xu, Hongwen Zhang, Lizhen Wang, Xiaochen Zhao, Han Huang, Guojun Qi,
  and Yebin Liu.
\newblock Latentavatar: Learning latent expression code for expressive neural
  head avatar.
\newblock In {\em ACM SIGGRAPH 2023 Conference Proceedings}, 2023.

\bibitem{Yu2021_BiSeNet}
Changqian Yu, Changxin Gao, Jingbo Wang, Gang Yu, Chunhua Shen, and Nong Sang.
\newblock Bisenet v2: Bilateral network with guided aggregation for real-time
  semantic segmentation.
\newblock {\em International Journal of Computer Vision}, 129:1--18, 11 2021.

\bibitem{Zakharov2019FewShotAL}
Egor Zakharov, Aliaksandra Shysheya, Egor Burkov, and Victor~S. Lempitsky.
\newblock Few-shot adversarial learning of realistic neural talking head
  models.
\newblock {\em Proceedings of IEEE International Conference on Computer
  Vision}, pages 9458--9467, 2019.

\bibitem{zeng23}
Bohan Zeng, Xuhui Liu, Sicheng Gao, Boyu Liu, Hong Li, Jianzhuang Liu, and
  Baochang Zhang.
\newblock Face animation with an attribute-guided diffusion model.
\newblock 2023.

\bibitem{zhang2022text2video}
Sibo Zhang, Jiahong Yuan, Miao Liao, and Liangjun Zhang.
\newblock Text2video: Text-driven talking-head video synthesis with
  personalized phoneme-pose dictionary.
\newblock In {\em IEEE International Conference on Acoustics, Speech and Signal
  Processing}, pages 2659--2663, 2022.

\bibitem{OneShotFace2019}
Yunxuan Zhang, Siwei Zhang, Yue He, Cheng Li, Chen~Change Loy, and Ziwei Liu.
\newblock One-shot face reenactment.
\newblock In {\em British Machine Vision Conference}, 2019.

\bibitem{Zheng22_imavatar}
Yufeng Zheng, Victoria~Fernández Abrevaya, Marcel~C. Bühler, Xu Chen,
  Michael~J. Black, and Otmar Hilliges.
\newblock {I} {M} {Avatar}: Implicit morphable head avatars from videos.
\newblock In {\em Proceedings of IEEE Conference on Computer Vision and Pattern
  Recognition}, 2022.

\bibitem{zhou2019talking}
Hang Zhou, Yu Liu, Ziwei Liu, Ping Luo, and Xiaogang Wang.
\newblock Talking face generation by adversarially disentangled audio-visual
  representation.
\newblock In {\em AAAI Conference on Artificial Intelligence}, 2019.

\bibitem{zhou2021cips}
Peng Zhou, Lingxi Xie, Bingbing Ni, and Qi Tian.
\newblock Cips-3d: A 3d-aware generator of gans based on
  conditionally-independent pixel synthesis.
\newblock {\em arXiv preprint arXiv:2110.09788}, 2021.

\bibitem{Yang2020_MakeItTalk}
Yang Zhou, Xintong Han, Eli Shechtman, Jose Echevarria, Evangelos Kalogerakis,
  and Dingzeyu Li.
\newblock Makeittalk: Speaker-aware talking-head animation.
\newblock {\em ACM Transactions on Graphics}, 39(6), 2020.

\bibitem{zhua23}
Yizhe Zhua, Chunhui Zhanga, Qiong Liub, and Xi Zhoub.
\newblock Audio-driven talking head video generation with diffusion model.
\newblock In {\em IEEE International Conference on Acoustics, Speech and Signal
  Processing}, pages 1--5, 2023.

\bibitem{zielonka2022towards}
Wojciech Zielonka, Timo Bolkart, and Justus Thies.
\newblock Towards metrical reconstruction of human faces.
\newblock In {\em European Conference on Computer Vision}, pages 250--269.
  Springer, 2022.

\bibitem{zielonka2023instant}
Wojciech Zielonka, Timo Bolkart, and Justus Thies.
\newblock Instant volumetric head avatars.
\newblock In {\em Proceedings of the IEEE/CVF Conference on Computer Vision and
  Pattern Recognition}, pages 4574--4584, 2023.

\end{thebibliography}
}

\end{document}